\documentclass[lettersize,journal]{IEEEtran}
\usepackage{amsmath,amsfonts}
\usepackage{algorithmic}
\usepackage{algorithm}
\usepackage{array}
\usepackage[caption=false,font=normalsize,labelfont=sf,textfont=sf]{subfig}
\usepackage{textcomp}
\usepackage{stfloats}
\usepackage{url}
\usepackage{verbatim}
\usepackage{graphicx}
\usepackage{cite}
\usepackage{booktabs}
\usepackage{multirow}
\usepackage{eccvabbrv}
\usepackage{xspace}

\def\method{Semantic Gaussians\xspace}

\begin{document}

\title{Semantic Gaussians: Open-Vocabulary Scene Understanding with 3D Gaussian Splatting}

\author{Jun Guo$^*$, Xiaojian Ma$^*$, Yue Fan, Huaping Liu$^\dag$,\IEEEmembership{Senior Member,~IEEE}, Qing Li$^\dag$

\thanks{* Equal contribution.}
\thanks{$\dag$ Corresponding author.}
\thanks{This paper was produced by BIGAI and Tsinghua University. They are in Beijing, China.}
}

\markboth{}%
{Guo \MakeLowercase{\textit{et al.}}: Semantic Gaussians: Open-Vocabulary Scene Understanding with 3D Gaussian Splatting}


\maketitle

\begin{figure*}[!t]
\begin{center}
    \includegraphics[width=0.9\linewidth]{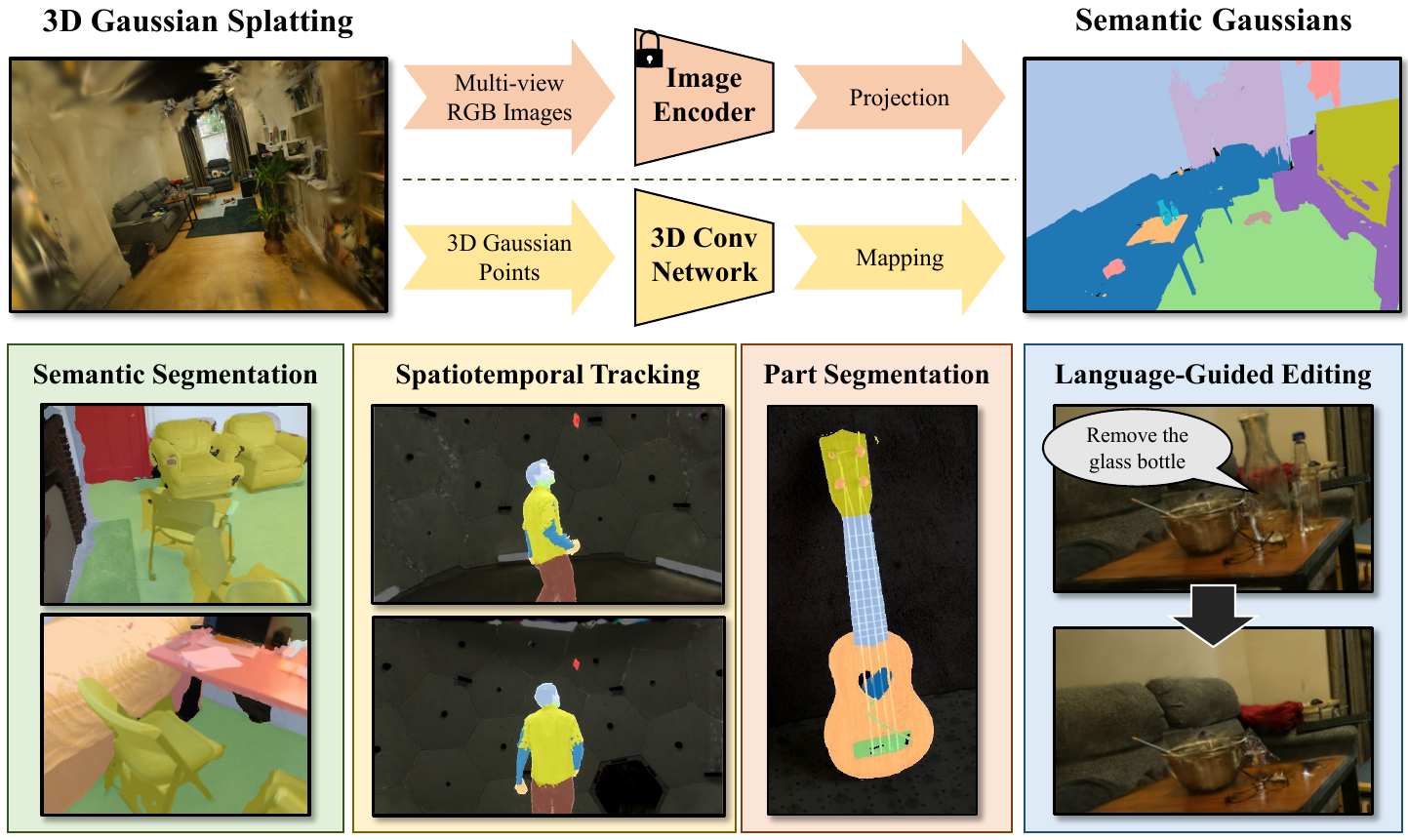}
\end{center}
\caption{Overview of our \method. We inject semantic features into off-the-shelf 3D Gaussian Splatting by either projecting semantic features from pre-trained 2D encoders or directly predicting pointwise embeddings by a 3D semantic network (or fusing these two). The newly added semantic components of 3D Gaussians open up diverse applications centered around open-vocabulary scene understanding.} 
\label{fig:teaser}
\end{figure*}

\begin{abstract}
    Open-vocabulary 3D scene understanding presents a significant challenge in computer vision, with wide-ranging applications in embodied agents and augmented reality systems.  Existing methods adopt neurel rendering methods as 3D representations and jointly optimize color and semantic features to achieve rendering and scene understanding simultaneously. In this paper, we introduce \method, a novel open-vocabulary scene understanding approach based on 3D Gaussian Splatting. Our key idea is to distill knowledge from 2D pre-trained models to 3D Gaussians. Unlike existing methods, we design a versatile projection approach that maps various 2D semantic features from pre-trained image encoders into a novel semantic component of 3D Gaussians, which is based on spatial relationship and need no additional training. We further build a 3D semantic network that directly predicts the semantic component from raw 3D Gaussians for fast inference. The quantitative results on ScanNet segmentation and LERF object localization demonstates the superior performance of our method. Additionally, we explore several applications of \method including object part segmentation, instance segmentation, scene editing, and spatiotemporal segmentation with better qualitative results over 2D and 3D baselines, highlighting its versatility and effectiveness on supporting diverse downstream tasks.
\end{abstract}


\section{Introduction}
\label{sec:intro}

Open-vocabulary 3D scene understanding is a crucial task in computer vision. Given a 3D scene, the goal is to comprehend and interpret 3D scenes with free-form natural language, \ie, without being limited to a predefined set of object categories. Allowing open-vocabulary scene queries enables machines to interact more effectively with the environment, facilitating tasks like object recognition, semantic scene reconstruction, and navigation in complex and diverse surroundings. Open-vocabulary 3D scene understanding has significant implications in various real-world applications such as robotics and augmented reality.

Various methods have been proposed to achieve open-vocabulary 3D scene understanding, relying on different 3D scene representations such as multi-view RGB images~\cite{rgb_segmentation}, point clouds~\cite{openscene,openmask3d}, and Neural Radiance Fields (NeRFs)~\cite{3dovs,lerf}. Approaches based on these representations have their pros and cons: Multi-view images are the most straightforward representation of 3D scenes, but allowing open-vocabulary understanding usually involves 2D vision-language models~\cite{alayrac2022flamingo}, which could struggle with consistency across different views, likely due to a lack of visual geometric knowledge; Point clouds are popular and well-studied, but the inherent sparsity nature of point clouds limits the application of open-vocabulary scene understanding upon them ~\cite{openscene}, \eg, it is challenging to obtain a dense prediction on a 2D view; 
injecting open-vocabulary semantics to NeRFs could enjoy both high 2D rendering quality and dense free-form visual recognition~\cite{lerf}, but the implicit design requires open-vocabulary recognition training for every new scene, and the rendering speed becomes a bottleneck to achieve real-time high-quality scene understanding.

A recent alternative scene representation is 3D Gaussian Splatting~(3DGS) proposed by Kerbl \etal~\cite{3dgs}. It utilizes 3D Gaussian points with color, opacity, and covariance matrix to represent the 3D scene, which can be learned from multi-view RGB images via gradient descent training. It attains the NeRF-level view rendering quality while preserving the explicit point-based characteristics similar to point clouds, making it suitable for open-vocabulary 3D scene understanding.


A main branch of previous approaches~\cite{dff, nerfsos, lerf, nfff, 3dovs, ovnerf, langsplat, feature3dgs, clipgs, fmgs, legaussians} is to adopt neural rendering methods like NeRF or 3DGS as the 3D representations, and jointly optimizing the color components and the semantic features, to achieve high-quality rendering and 3D scene understanding from arbitrary 2D views. The semantic knowledge is usually distilled from open-vocabulary 2D foundation models, such as CLIP~\cite{clip} or LSeg~\cite{lseg}, whose outputs predicted on training views serve as weak supervision during optimization.

In this work, we propose \method, a novel approach to open-vocabulary 3D scene understanding building upon the benefits of 3D Gaussian Splatting. The core idea of \method is to distill the knowledge from pre-trained 2D encoders into 3D Gaussians, thereby assigning a semantic component to each Gaussian point. To achieve this, we establish correspondence between 2D pixels and 3D Gaussian points and propose a versatile projection framework to map the semantic features of 2D pixels onto each 3D Gaussian point. Our framework is rather flexible and can leverage arbitrary pre-trained 2D models, such as OpenSeg~\cite{openseg}, CLIP~\cite{clip}, VLPart~\cite{vlpart}, \etc, to generate pixel-wise semantic features on 2D RGB images. Compared to previous approachs, our method injects semantic components into 3D Gaussians \textit{without additional training}, allowing for effective open-vocabulary scene queries.

In addition to projection, we further introduce a 3D semantic network that directly predicts open-vocabulary semantic components out of raw 3D Gaussians. Specifically, we employ MinkowskiNet~\cite{minknet}, a 3D sparse convolution network to process 3D Gaussians. The 3D convolution network takes raw RGB Gaussians as input and is supervised by the semantic components of Gaussians obtained from the aforementioned projection method. As a result, we may simply run this network to obtain the semantic components, enabling faster inference. This network leverages geometric attributes to understand unseen scenes, boosting the generalizability and robustness of our method beyond 2D projection. Note that the prediction of the 3D semantic network can be combined with the projected features to further improve the quality of semantic components in Gaussians and open-vocabulary scene understanding performances.

We conduct experiments on the ScanNet semantic segmentation benchmark~\cite{scannet} and LERF localization~\cite{lerf}, and prove our efficiency compared to 2D pre-trained models. Besides segmentation and localization, we also explore diverse applications of \method, including 3D part segmentation on the MVImgNet object dataset~\cite{mvimgnet}, instance segmentation and scene editing in multi-object scenes, and spatiotemporal tracking on 4D dynamic Gaussians~\cite{dynamic3dgaussians}. 


In summary, our contributions are three-fold: 
\begin{enumerate}
    \item We introduce \method, a novel approach to open-vocabulary 3D scene understanding by bringing a novel semantic component to 3D Gaussian Splatting.    
    \item We propose a versatile semantic feature projection framework to map various pre-trained 2D features to 3D Gaussian points, and introduce a 3D semantic network to further allow direct prediction of these semantic components from raw 3D Gaussians;
    \item We conduct experiments on the ScanNet and LERF localization datasets to demonstrate the effectiveness of our method on open-vocabulary scene understanding and explore various applications including object part segmentation, instance segmentation, scene editing, and spatiotemporal tracking.
\end{enumerate}

\section{Related Work}
\label{sec:related_work}

\subsection{3D Scene Representation}

Modeling and representing 3D scenes are crucial initial steps in understanding such environments. Before the advent of deep learning, common methods involved simplifying scenes into combinations of basic elements. Classic approaches include point clouds, meshes, and voxels. Point clouds represent scenes as collections of points, where each point's XYZ coordinates together form the scene's geometric shape. Enhancing point clouds with RGB values, semantic labels, and other data enriches their ability to represent scenes. Meshes depict 3D scene surfaces using collections of polygons, with triangular meshes being the most common, representing surfaces as interconnected triangles. By recording the vertices and adjacency relationships of each triangle, 3D scenes can be effectively represented. Voxels discretize continuous 3D space into cubic units, extending the concept of pixels from 2D to 3D. Although these traditional methods have seen success, their limitations are increasingly apparent in today's pursuit of realistic reconstruction and rendering.

On the other hand, implicit neural representations, represented by Neural Radiance Fields (NeRF), have made remarkable strides in various 3D computer vision tasks. NeRF was initially proposed by Mildenhall \etal~\cite{nerf} to address the problem of novel view synthesis. By extracting shape and color information from images captured from multiple viewpoints and learning a continuous 3D radiance field via neural networks, NeRF achieves photorealistic rendering of 3D scenes from arbitrary viewpoints and distances. Succeeding works demonstrate the capability of NeRF in 3D scene representation. Semantic-NeRF~\cite{semanticnerf} explored encoding semantics into a NeRF to achieve 3D scene understanding. EditNeRF~\cite{editnerf} defines a conditional NeRF where 3D objects are conditioned on shape and appearance codes to achieve scene editing. Some works~\cite{dynamicnerf,dnerf} jointly predict a canonical space and a temporal deformation field to achieve dynamic scene reconstruction. 


Recently, Kerbl \etal~\cite{3dgs} have proposed a new novel view synthesis method called 3D Gaussian Splatting, which represents the 3D scene with a set of 3D Gaussians. This method has demonstrated real-time rendering capabilities at 1080p resolution, achieving a remarkable 60 frames per second while maintaining state-of-the-art visual quality. The innovation behind 3D Gaussian Splatting lies in its incorporation of point-based $\alpha$-blending and a differentiable tile rasterizer, enabling efficient rendering. Though it obtains a dense set of Gaussians via optimization, all parameters in these 3D Gaussians are explicit and editable.  The speed enhancement and explicit parameterization position 3D Gaussian Splatting as a highly promising representation method. Building upon its success in novel view synthesis, some studies~\cite{dynamic3dgaussians, 4dgs, 4dgs_2, 4dgs_3, 4dgs_gen, 4dgs_gen2} have extended 3DGS to dynamic scenes. For example, Luiten \etal~\cite{dynamic3dgaussians} extended the concept to Dynamic 3D Gaussians, explicitly modeling 3D Gaussians at different time steps to accommodate 4D dynamic scenes. Furthermore, many recent works~\cite{textto3d,generation1,generation2,generation3,generation4,4dgs_gen,4dgs_gen2} leverage 3D Gaussian Splatting to achieve high-quality text-to-3D or image-to-3D generation. In this study, we propose an open-vocabulary 3D scene understanding method, leveraging the advantages offered by 3D Gaussian Splatting.

\subsection{Open-Vocabulary Scene Understanding}

\noindent
\subsubsection{Scene understanding from 2D}
Encouraged by the availability of adequate text-image datasets and the advancement in vision language models, the field of 2D open-vocabulary scene understanding has made significant progress in recent years. Prevailing approaches~\cite{ovseg,segclip,lseg} distill knowledge from large-scale pre-trained foundation models (\eg, CLIP~\cite{clip}) to achieve zero-shot understanding, including recognizing long-tail objects and understanding synonymous labels. However, these methods are limited to small partial scenes represented by a single 2D image. When it comes to 3D scenes, the prediction result of these 2D models can hardly remain consistent between different angles of view. In contrast, our work relies on these 2D pre-trained models to achieve 3D scene understanding, segmenting, and understanding the scene from a panoptic perspective. Moreover, the proposed 3D network can perform 3D-only scene understanding in the absence of 2D pre-trained models and images.

\noindent
\subsubsection{Scene understanding from 3D}
Open-vocabulary 3D scene understanding has been a long-standing challenge in computer vision. Some point-cloud-based methods\cite{openscene, conceptfusion, pla, clipfo3d} encode the semantic features from 2D pre-trained models into 3D scene points to achieve open-vocabulary 3D scene understanding. To achieve high-quality rendering and scene understanding simultaneously, feature field distillation in NeRF has been well explored. Early works such as Semantic-NeRF~\cite{semanticnerf}, Panoptic Lifting~\cite{panopticlifting} and Contrastive Lift~\cite{contrastivelift} embed semantic labels into NeRF, resulting in precise 3D segmentation maps. Encouraged by this idea, another branch of methods~\cite{dff, nerfsos, lerf, nfff, 3dovs, ovnerf} integrate semantic embeddings from pre-trained models such as LSeg~\cite{lseg}, CLIP or DINO~\cite{dino} into NeRFs, achieving open-vocabulary 3D scene understanding. Recently, some works~\cite{langsplat, feature3dgs, clipgs, fmgs, legaussians} have made efforts to transfer those NeRF-based methods to 3DGS, obtaining 3DGS with semantic features via optimization. Our work shares a similar idea with those methods, while the \method requires no extra training for Gaussians. The explicit nature of 3D Gaussian Splatting enables \method to achieve versatile projection from 2D semantic maps into 3D Gaussian points.



\section{\method}
\label{sec:method}

\begin{figure*}[t!]
\begin{center}
    \includegraphics[width=\linewidth]{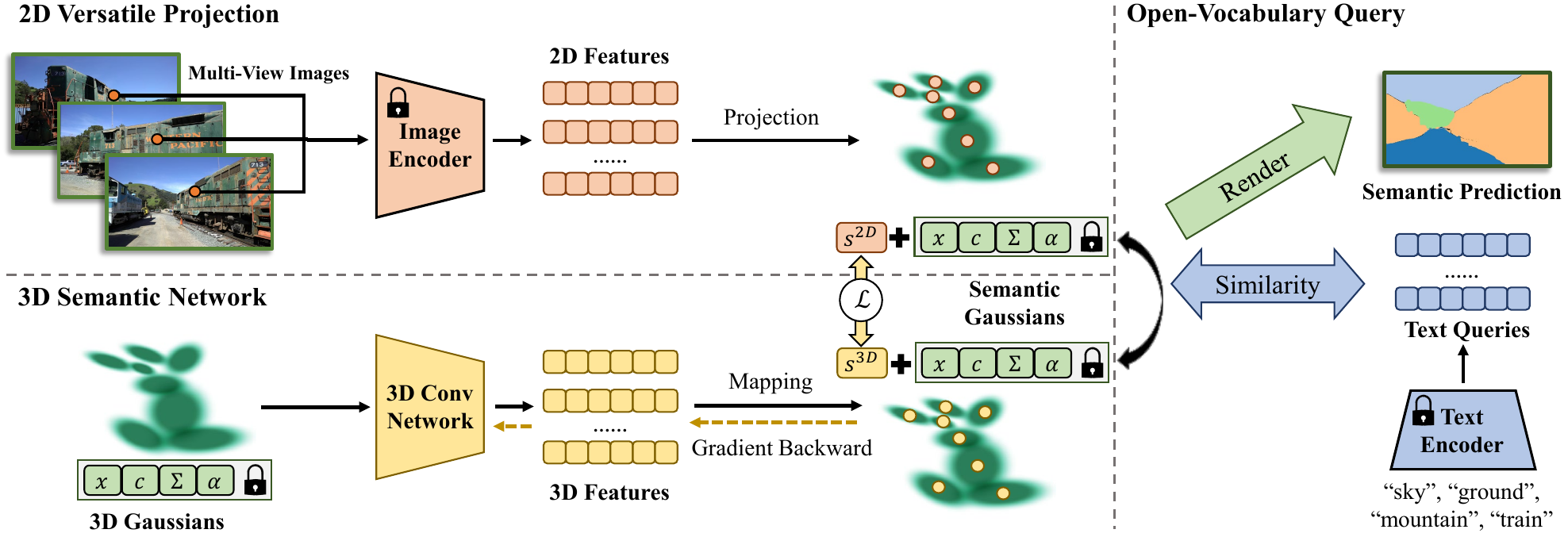}
\end{center}
\caption{An illustration of the pipeline of \method. \textit{Upper left:} our projection framework maps various pre-trained 2D features to the semantic component $s^{\text{2D}}$ of 3D Gaussians; \textit{Bottom left:} we additionally introduce a 3D semantic network that directly predicts the semantic components $s^{\text{3D}}$ out of raw 3D Gaussians. It is supervised by the projected $s^{\text{2D}}$; \textit{Right:} given an open-vocabulary text query, we compare its embedding against the semantic components ($s^{\text{2D}}$, $s^{\text{3D}}$, or their fusion) of 3D Gaussians. The matched Gaussians will be splatted to render the 2D mask corresponding to the query.} 
\label{fig:framework}
\end{figure*}

In this section, we illustrate the framework of our \method. Fig.~\ref{fig:framework} depicts the overall framework of our method. \method starts from a group of 3D Gaussians (Sec.~\ref{sec:3dgs}), performing scene understanding on it through 2D versatile projection and 3D semantic network processing. We first introduce our versatile projection method that projects 2D semantic embeddings from various pre-trained vision-language models into 3D Gaussian points (Sec.~\ref{sec:2d_proj}). We then depict the 3D semantic network that learns from the projected features and predicts the semantics of 3D Gaussians in unseen scenes (Sec.~\ref{sec:3d_network}). At last, we describe the feature ensemble process and use the prediction result to support various applications (Sec.~\ref{sec:inference}).

\subsection{3D Gaussian Splatting}
\label{sec:3dgs}

To achieve general 3D open-vocabulary scene understanding, we employ 3D Gaussian Splatting~\cite{3dgs} as the representation of 3D scenes. 3DGS can render images from arbitrary viewpoints in a differentiable manner, thus effectively leveraging the knowledge of various 2D foundation models. Specifically, we achieve scene understanding by rendering 2D semantic images from specified viewpoints with 3DGS.

3DGS consists of a set of learnable 3D Gaussian points, where each point has a 3D coordinate $\mu$ representing its position, a covariance matrix $\Sigma$ representing its shape, spherical harmonic parameters $c$ representing its color, and an opacity value $\alpha$ representing its transparency. 3DGS can be constructed from multi-view images and can utilize information from Structure-from-Motion (SfM) point clouds~\cite{sfm} for initialization, thereby achieving better rendering quality and geometric structure.

3DGS uses point-based $\alpha$-blending to compute pixel values on 2D images. The value of each pixel $C$ is given by volumetric rendering along a ray:
\begin{equation}
    C = \sum_{i\in \mathcal{N}} c_i \alpha_i T_i \text{ with } T_i=\prod_{j=1}^{i-1}(1-\alpha_j),
\end{equation}
where $\mathcal{N}$ is the set of sorted Gaussians in front-to-back depth order overlapping with the given pixel. $\alpha_i$ is the opacity of point $i$, and $c_i$ is the color of point $i$, which is calculated by spherical harmonics.

To project 3D Gaussians onto a certain 2D plane, Zwicker et al. proposed a splatting method to calculate the covariance matrix $\Sigma'$ from the camera's viewpoint. Given the world-to-camera transformation matrix $W$, the covariance matrix in camera coordinates is given as follows:
\begin{equation}
    \Sigma'=JW\Sigma W^T J^T,
\end{equation}
where J is the Jacobian of the affine approximation of the projective transformation. If we skip the third row and column of $\Sigma'$, we can obtain the 2D variance matrix. To assure the positive semi-definite, the covariance matrix $\Sigma$ is decomposed into rotation matrix $R$ and scaling matrix $S$, which can be represented as follows:
\begin{equation}
    \Sigma = RSS^T R^T.
\end{equation}

3DGS can be regarded as a special point cloud with additional features, thus sharing some properties of point clouds. An intuitive idea is to project the semantic information obtained from 2D foundation models onto the corresponding Gaussian points based on spatial relationships, rather than through differentiable rasterization and rendering. When rendering semantic maps, it is sufficient to consider the correspondence of geometric positions without accounting for complex lighting conditions. Based on this idea, we propose \method to achieve versatile scene understanding.

\subsection{2D Versatile Projection}
\label{sec:2d_proj}
The first contribution of our approach is a versatile feature projection method. We extract pixel-level semantic maps for RGB images from a 2D pre-trained model and project them into 3D Gaussians of a scene.

\subsubsection{Semantic Map Extraction}
Our method starts from off-the-shelf 3D Gaussians $\mathbf{G}$ of a certain scene. We can use ground-truth RGB images that are used to train 3D Gaussians or render RGB frames via the 3D Gaussians. Owing to the photorealistic rendering performance of 3D Gaussian Splatting, \method can run without 2D ground-truth images. Given RGB images $\mathbf{I}$ with a shape of $H\times W$, the aim of \method is to get pixel-level semantic maps denoted by $s \in \mathbb{R}^{H\times W \times C}$ from an arbitrary 2D vision-language model $\mathcal{E}^\text{2D}$. The most straightforward avenue to obtain per-pixel semantic maps is utilizing pixel-level segmentation models such as OpenSeg. However, leveraging other encoders, \eg VLPart~\cite{vlpart} could help with object parts semantics (not covered by OpenSeg). Moreover, features from different types of models can be integrated as an ensemble to produce more accurate results. Therefore, our versatile projection method should be able to reconcile various visual features.

\subsubsection{Unifiying Various 2D Features with SAM}
Accommodating a variety of 2D pre-trained features is \textbf{non-trivial} as they can be pixel-level segmentation network (\eg, OpenSeg~\cite{openseg}, LSeg~\cite{lseg}), instance-level recognition network (\eg, GroundingDINO~\cite{groundingdino}, VLPart~\cite{vlpart}), or image-level classification network (\eg, CLIP~\cite{clip}). Additionally, \method are able to utilizes Segment Anything (SAM)~\cite{sam} to produce fine segmentation maps for each model. 
For pixel-level models, SAM can refine the segmentation boundary. Given an RGB image $\mathbf{I}$, we use everything prompt in SAM to generate $N$ binary masks $\mathbf{M}_1, \cdots, \mathbf{M}_N$. We calculate the average pooling of embeddings in each mask $\mathbf{M}_i$, and assign it as the embedding of all pixels in this mask: $s[\mathbf{M}_i] = AvgPool(s[\mathbf{M}_i])$.
For instance-level models, we use SAM as the postprocessing module to get fine masks. Similar to Grounded-SAM~\cite{groundedsam}, we use the prediction result from the pre-trained model as the box prompt of SAM. After getting the binary mask, we assign the CLIP embedding of this instance to all the pixels within the instance region.
For image-level models, SAM can be a preprocessing module to get region proposals. We use the ``everything'' prompt in SAM to get various proposed regions. Each region is padded, cropped, and resized to $224\times 224$, and fed into the image-level model to get semantic embeddings. Similarly, the semantic embeddings are assigned to all pixels within each proposed region.


\subsubsection{2D-3D Projection and Fusion}
After acquiring per-pixel semantic maps, \method projects them into 3D Gaussians to obtain the semantic components. For each pixel $\mathbf{u}=(u, v)$ in a semantic mapping $s$, \method tries to find if there are corresponding 3D Gaussian points $\mathbf{p}=(x,y,z)$ in the space. This can be achieved when the camera intrinsic matrix $K$ and world-to-camera extrinsic matrix $E$ are provided. Under the pinhole camera model, the projection can be formulated as $\Tilde{\mathbf{u}} = K\cdot E \cdot \Tilde{\mathbf{p}}$, where $\Tilde{\mathbf{u}}$ and $\Tilde{\mathbf{p}}$ are the homogeneous coordinates of $\mathbf{u}$ and $\mathbf{p}$. 
After this projection, every pixel $\mathbf{u}$ will correspond to a beam of ray in the 3D space. As we only expect to project 2D semantics to the surface points in 3D space, we perform depth rendering of 3D Gaussians to get the depth map of the 3D scene. During splatting and volume rendering, the opacity is accumulated from near to far. Therefore, we set an opacity threshold $\alpha_d$, and when the opacity surpasses $\alpha_d$, the ray of view is occluded by some opaque objects, where we can record the depth. Note that we do not need ground-truth depth from datasets.

When 2D pixels and 3D Gaussian points are paired, assuming that a certain Gaussian point $\mathbf{p}$ in 3D spaces has a group of 2D semantics $\{s_1, \cdots, s_K\}$ from $K$ different views, these semantics can be fused by average pooling: $s^{\text{2D}}_\mathbf{p}=AvgPool(s_1, \cdots, s_K)$. By repeating this process for all 3D Gaussian points, we can construct a group of semantic Gaussians for a 3D scene.

\begin{figure*}[!ht]
\begin{center}
    \includegraphics[width=\linewidth]{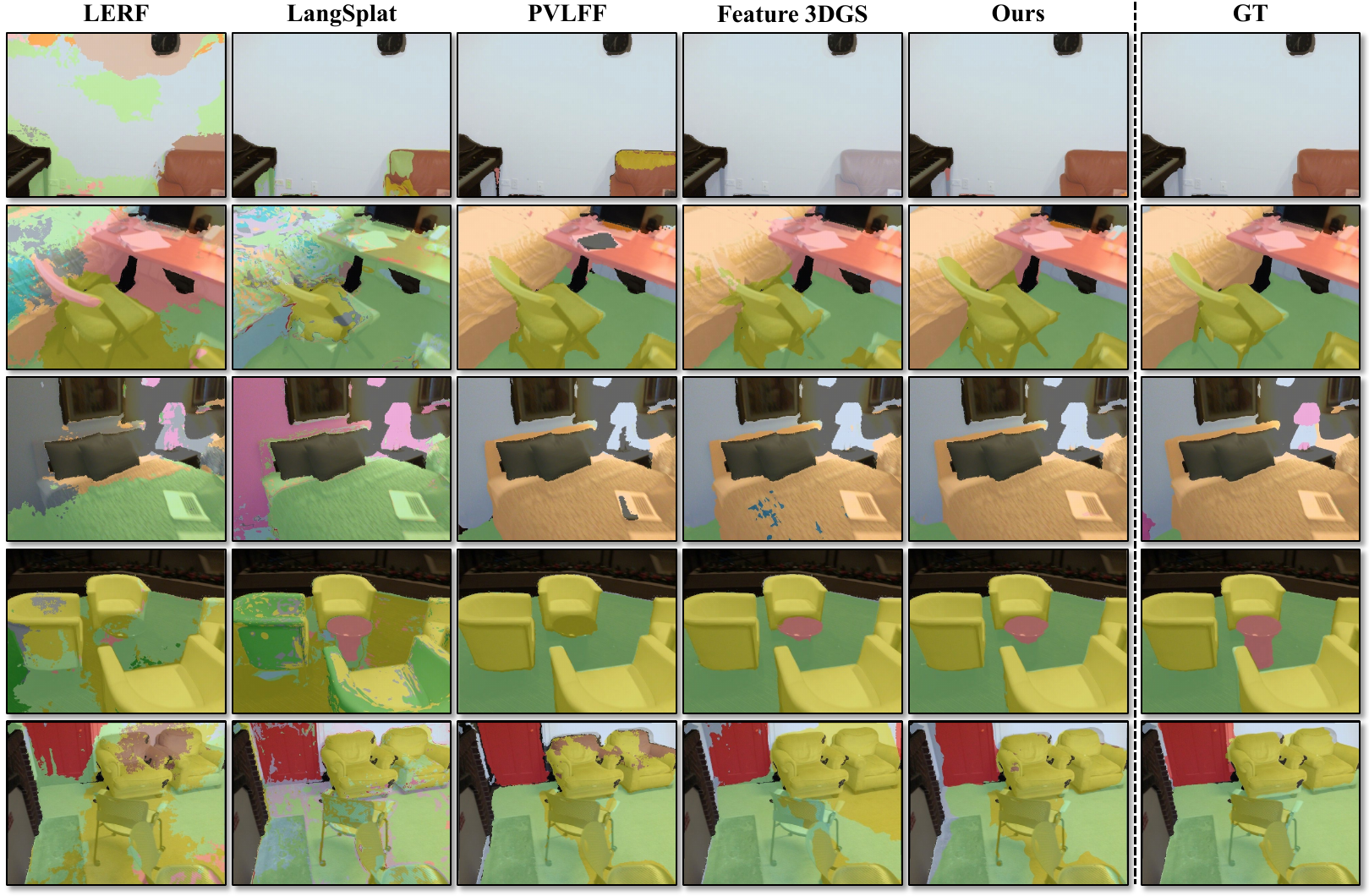}
    \includegraphics[width=\linewidth]{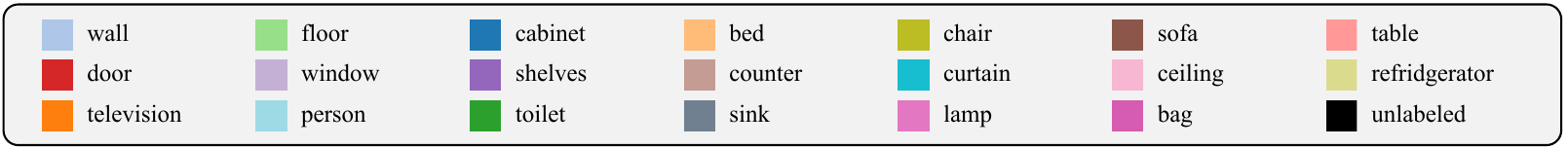}
\end{center}
\caption{Visualization of scene-level semantic segmentation performance for open-vocabulary 3D scene understanding methods on ScanNet dataset.}
\label{fig:sem_seg}
\end{figure*}

\subsection{3D Semantic Network}
\label{sec:3d_network}
In addition to projecting 2D pre-trained features onto 3D Gaussians, we alternatively explore a more direct approach -- predicting the semantic components from the raw 3D Gaussians. In this section, we build a 3D semantic network $f^\text{3D}$ to do exactly that.
Specifically, given the input 3D Gaussians $\mathbf{G}$, our 3D network $f^\text{3D}$ predict the point-wise semantic component $s^\text{3D}$, which can be formulated as Eqn.~\ref{eqn:3d_model}:

\begin{equation}
\label{eqn:3d_model}
    s^\text{3D} = f^\text{3D}(\mathbf{G}).
\end{equation}

We use fused features from Sec.~\ref{sec:2d_proj} (denoted by $s^\text{2D}$) to supervise the 3D model. The loss function is the cosine similarity loss:

\begin{equation}
\label{eqn:loss}
    \mathcal{L} = 1 - \cos(s^\text{3D}, s^\text{2D}).
\end{equation}

We use MinkowskiNet~\cite{minknet} as the backbone of our 3D model. MinkowskiNet is a 3D sparse convolution network designed for point clouds. Due to the similarity between 3D point clouds and 3D Gaussians, it can be utilized to process 3D Gaussians. Opacities, colors, and covariance matrixes are set as input features of our 3D model, and the output is the semantic embeddings $s^\text{3D}$ of every Gaussian point.

Though the supervised target entirely comes from pre-trained 2D encoders, the 3D model recognizes the scene by processing 3D geometric information rather than multiple 2D views, making the result more consistent. Experiments in Sec.~\ref{sec:sem_seg} show that the 3D prediction $s^{\text{3D}}$ can complement 2D projection features $s^{\text{2D}}$. Also, the inference speed of our 3D semantic model is much faster than 2D projection.

\subsection{Inference}
\label{sec:inference}
After obtaining the semantic components of 3D Gaussians, we can perform language-driven open-vocabulary scene understanding. In this section, we will detail the inference process. Given a free-form language query, we use the CLIP text encoder to encode the prompts into text embedding $\mathbf{t}$. We calculate the cosine similarity between the text embedding and the semantic component of every 3D Gaussian point and the matched Gaussians will be viewed as corresponding to the query. Take semantic segmentation and part segmentation as an example, labels of $N$ semantic classes are encoded as $\mathbf{t}_1, \cdots, \mathbf{t}_N$. We calculate the cosine similarity between these text embeddings and the semantic embedding of 3D Gaussians:

\begin{equation}
    c_n^\text{2D} = 1 - cos(s^\text{2D}, \mathbf{t}_n), c_n^\text{3D} = 1 - cos(s^\text{3D}, \mathbf{t}_n)
\end{equation}

When 2D and 3D semantic components $s^\text{2D}$ and $s^\text{3D}$ both exists, we choose the larger value in $c_n^\text{2D}$ and $c_n^\text{3D}$ as the cosine similarity of the class $\mathbf{t}_n$. The similarities $\{c_1, \cdots, c_n\}$ after the softmax function can be the confidence score of each class. To get a 2D semantic segmentation map, the confidence scores are splatted onto 2D views, which is similar to RGB splatting.

\section{Experiments}
\label{sec:experiments}
In this section, we conduct various experiments to demonstrate the effectiveness of \method on open-vocabulary 3D scene understanding and other applications. We first evaluate and compare our method on the ScanNet 2D semantic segmentation benchmark which is constructed on 3D scenes. Then, we exhibit the 3D object localization results of \method on LERF dataset. Next, we exhibit qualitative results on some applications, including part segmentation, spatiotemporal tracking, and language-guided editing. We also provide ablation study results.

\subsection{Experimental Setup}

\subsubsection{Datasets}
For comparisons on scene-level semantic segmentation, we select ScanNet~\cite{scannet} dataset as our 2D segmentation benchmark. ScanNet is a large-scale segmentation benchmark of indoor scenes, with calibrated RGBD trajectories, 3D point clouds, and ground-truth semantic label maps. We train 3D RGB Gaussians for all 1,201 scenes. We train our 3D semantic network on ScanNet train set and evaluate our performance on 12 scenes~\cite{panopticlifting} from the validation set. To compare with closed-set methods, we follow the setting of \cite{panopticlifting} which maps ScanNet-20 classes to 21 classes from COCO dataset. 

For 3D object localization tasks, we choose LERF~\cite{lerf} dataset as our benchmark. LERF dataset provide several 3D scenes containing long-tail objects and multi-scale semantics. The dataset is captured by iPhone App Polycam to get multi-view images and SfM points. In our experiments, we follow the setting of \cite{langsplat} to evaluate the localization accuracy on 4 different scenes.

For qualitative evaluations, we choose MVImgNet~\cite{mvimgnet} dataset as our part segmentation dataset, and CMU Panoptic~\cite{cmupanoptic} dataset as our spatiotemporal tracking dataset. MVImgNet is a multi-view single-object dataset containing 238 classes of objects with camera parameters and sparse point clouds. CMU Panoptic dataset is a large-scale dataset for multi-people engaging in social activities. For language-guided editing, we choose some scenes in the Mip-NeRF 360~\cite{mipnerf360} dataset to show our performance.

\subsubsection{Implementation Details}

All our experiments are trained on a NVIDIA RTX 4090 GPU. We train 10000 iterations for RGB Gaussians and 100 epochs for 3D semantic network. For scene-level semantic segmentation experiments, We apply LSeg to generate pixel-level open-vocabulary semantic features for 2D projection. As for 3D semantic network, we use MinkowskiNet34A~\cite{minknet} as our backbone. For 3D object localization tasks, we both apply CLIP with SAM and LSeg as our 2D pretrained model.

For part segmentation and spatiotemporal tracking, we use VLPart as our projection model. To evaluate our \method on 4D Gaussians, we follow the work of Dynamic 3D Gaussians~\cite{dynamic3dgaussians} to obtain dynamic Gaussians with temporal information.

\subsection{Quantitative Results}

\begin{table}[!ht]
\caption{2D semantic segmentation results on 12 scenes in ScanNet validation set. We report the mean IoU and the mean accuracy on all classes.}
\label{tab:2d_seg}
\setlength{\tabcolsep}{12pt}
\centering
\begin{tabular}{@{}lccc@{}}
\toprule
\ \ Method                           & Backbone             & mIoU                 & mAcc                 \\ \midrule
\ \ \textit{closed-set methods}      & \multicolumn{1}{l}{} & \multicolumn{1}{l}{} & \multicolumn{1}{l}{} \\
\ \ Mask2Former~\cite{mask2former}   & ViT                  & 46.7                 & -                    \\
\ \ DM-NeRF~\cite{dmnerf}            & NeRF                 & 49.5                 & -                    \\
\ \ SemanticNeRF~\cite{semanticnerf} & NeRF                 & 59.2                 & -                    \\
\ \ Panoptic Lifting~\cite{panopticlifting} & NeRF                 & \textbf{65.2}        & -                    \\ \midrule
\ \ \textit{open-vocabulary methods} & \multicolumn{1}{l}{} & \multicolumn{1}{l}{} & \multicolumn{1}{l}{} \\
\ \ OpenSeg~\cite{openseg}           & EfficientNet         & 53.4                 & 75.1                 \\
\ \ LSeg~\cite{lseg}                 & ViT                  & 56.1                 & 74.5                 \\
\ \ LERF~\cite{lerf}                 & NeRF+CLIP            & 31.2                 & 61.7                 \\
\ \ PVLFF\cite{pvlff}                & NeRF+LSeg            & 52.9                 & 67.0                 \\
\ \ LangSplat~\cite{langsplat}       & 3DGS+CLIP            & 24.7                 & 42.0                 \\
\ \ Feature3DGS~\cite{feature3dgs}   & 3DGS+LSeg            & 59.2                 & 75.1                 \\ \midrule
\ \ Ours 2D                          & 3DGS+LSeg            & 61.0                 & 76.6                 \\
\ \ Ours 3D                          & 3DGS+LSeg            & 59.7                 & 74.7                 \\
\ \ Ours 2D+3D                       & 3DGS+LSeg            & \textbf{62.0}        & \textbf{77.0}        \\ \bottomrule
\end{tabular}
\end{table}

\begin{table}[!ht]
\caption{3D object localization results on LERF dataset. We follow LangSplat~\cite{langsplat} to report localization accuracy (\%) on 4 scenes.}
\label{tab:localization}
\setlength{\tabcolsep}{4pt}
\centering
\begin{tabular}{@{}lccccc@{}}
\toprule
\ \ Scene                     & LSeg & LERF & LangSplat     & Ours (LSeg) & Ours (CLIP) \\ \midrule
\ \ \textit{ramen}            & 14.1 & 62.0 & 73.2          & 21.1        & \textbf{76.8}   \\
\ \ \textit{figurines}        & 8.9  & 75.0 & 80.4          & 10.7        & \textbf{83.1}   \\
\ \ \textit{teatime}          & 33.9 & 84.8 & 88.1          & 32.2        & \textbf{89.8}   \\
\ \ \textit{waldo\_kitchen}   & 27.3 & 72.7 & \textbf{95.5} & 31.8        & 90.9            \\ \midrule
\ \ overall                   & 21.1 & 73.6 & 84.3          & 24.0        & \textbf{85.2}   \\ \bottomrule
\end{tabular}
\end{table}

\begin{table}[!ht]
\caption{Performance of ablation studies.}
\label{tab:ablation}
\setlength{\tabcolsep}{14pt}
\centering
\begin{tabular}{@{}lcc@{}}
\toprule
\ \ Setting              & mIoU & mAcc \\ \midrule
\ \ original             & 62.0 & 77.0 \\
\ \ XYZ+RGB features     & 58.9 & 74.7 \\
\ \ 20\% Gaussian points & 61.2 & 76.1 \\
\ \ 10\% input views     & 59.8 & 75.2 \\ \bottomrule
\end{tabular}
\end{table}

\subsubsection{Open-Vocabulary Semantic Segmentation}
\label{sec:sem_seg}

\begin{figure*}[!ht]
\begin{center}
    \subfloat[chair]{\includegraphics[width=0.32\linewidth]{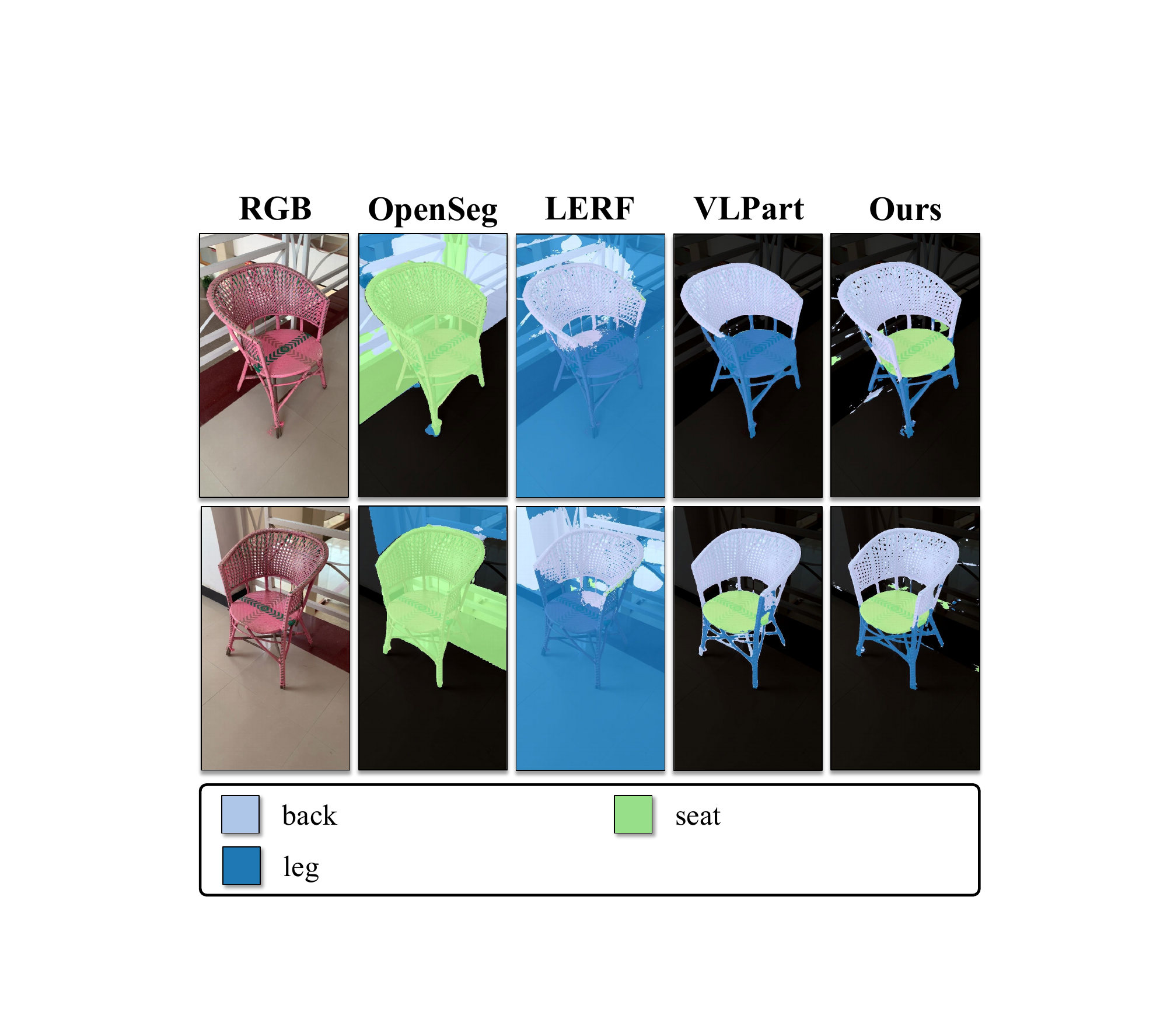}}
    \subfloat[scissors]{\includegraphics[width=0.32\linewidth]{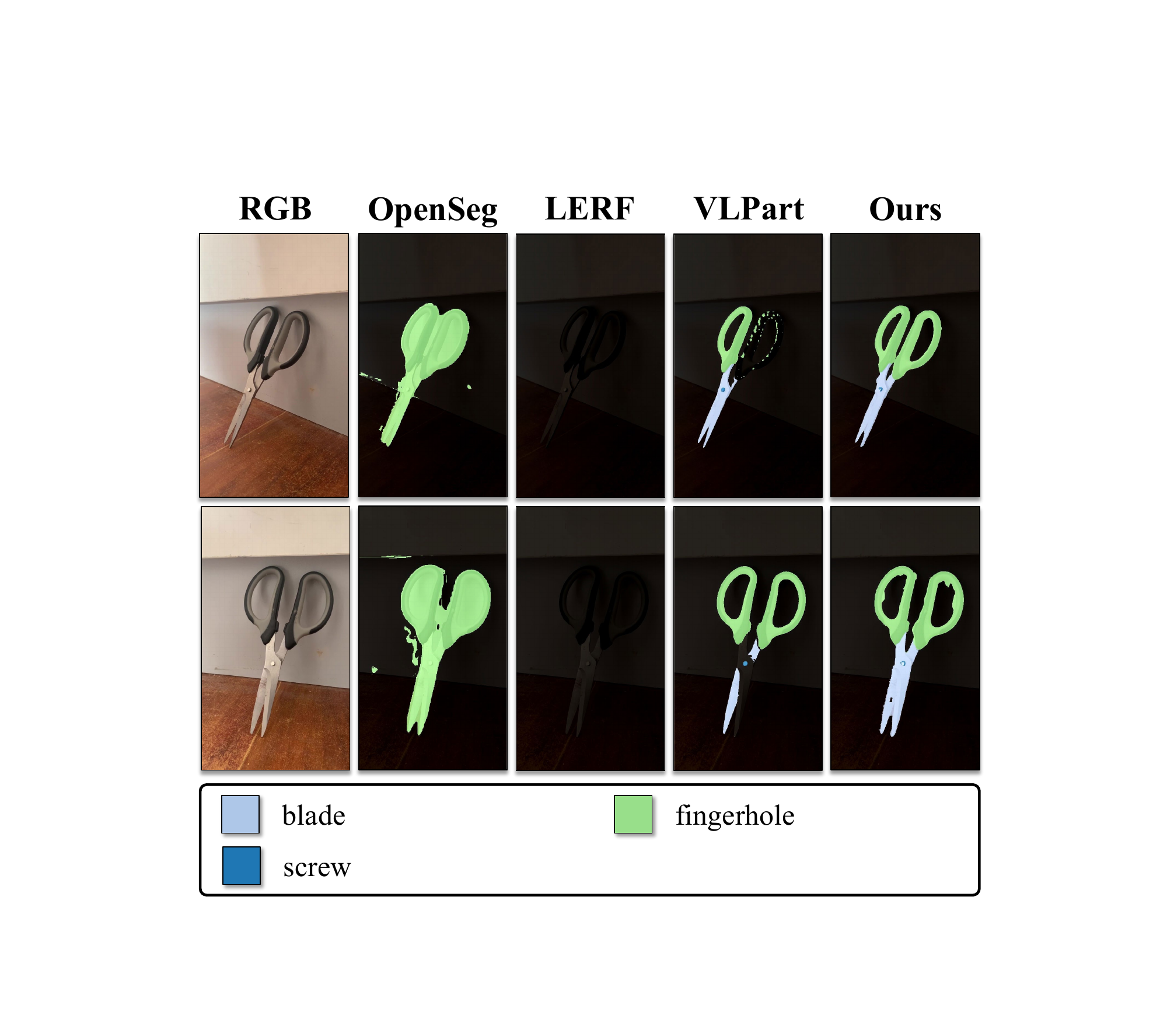}}
    \subfloat[basket]{\includegraphics[width=0.32\linewidth]{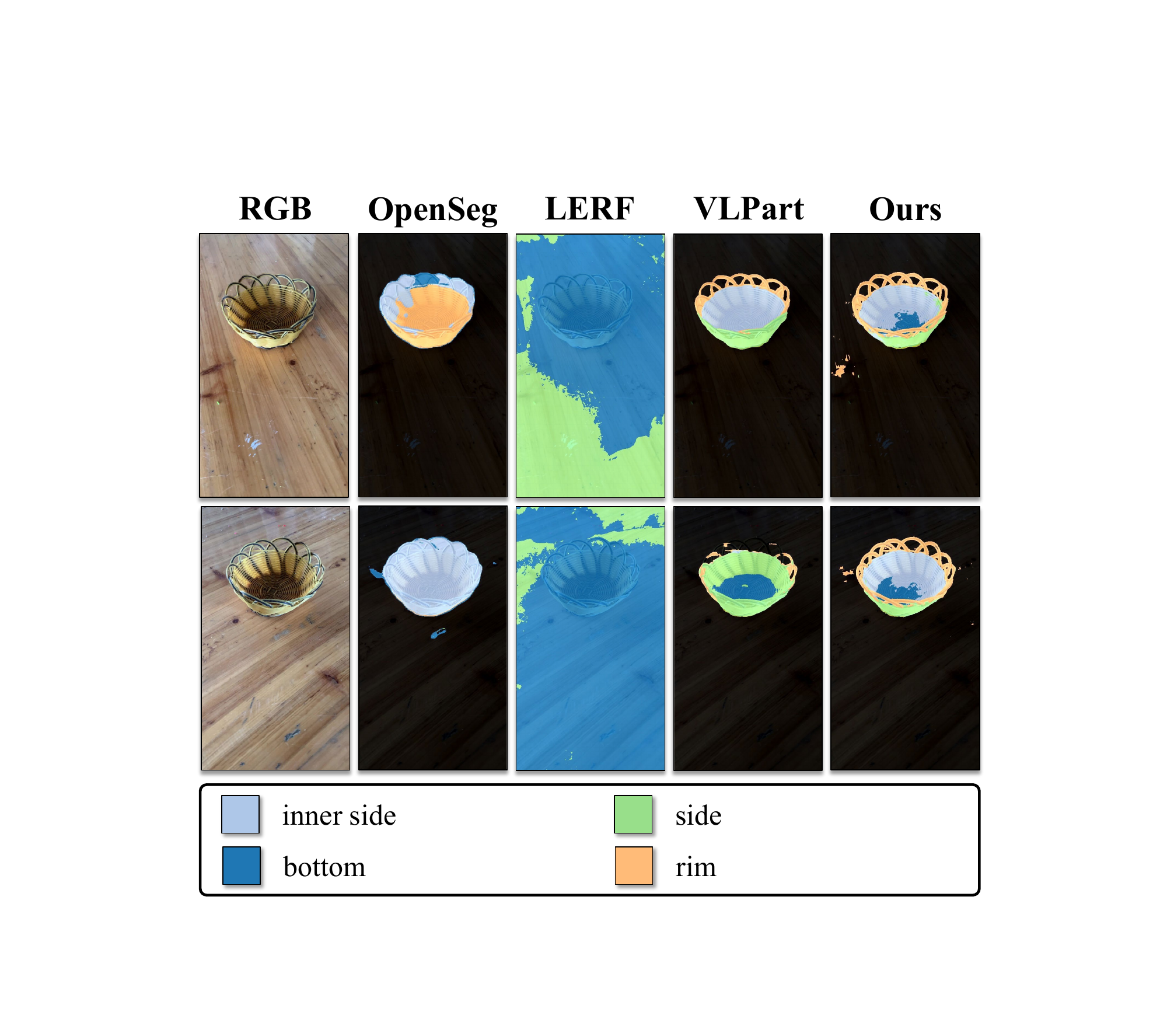}} \\
    \subfloat[shoe]{\includegraphics[width=0.32\linewidth]{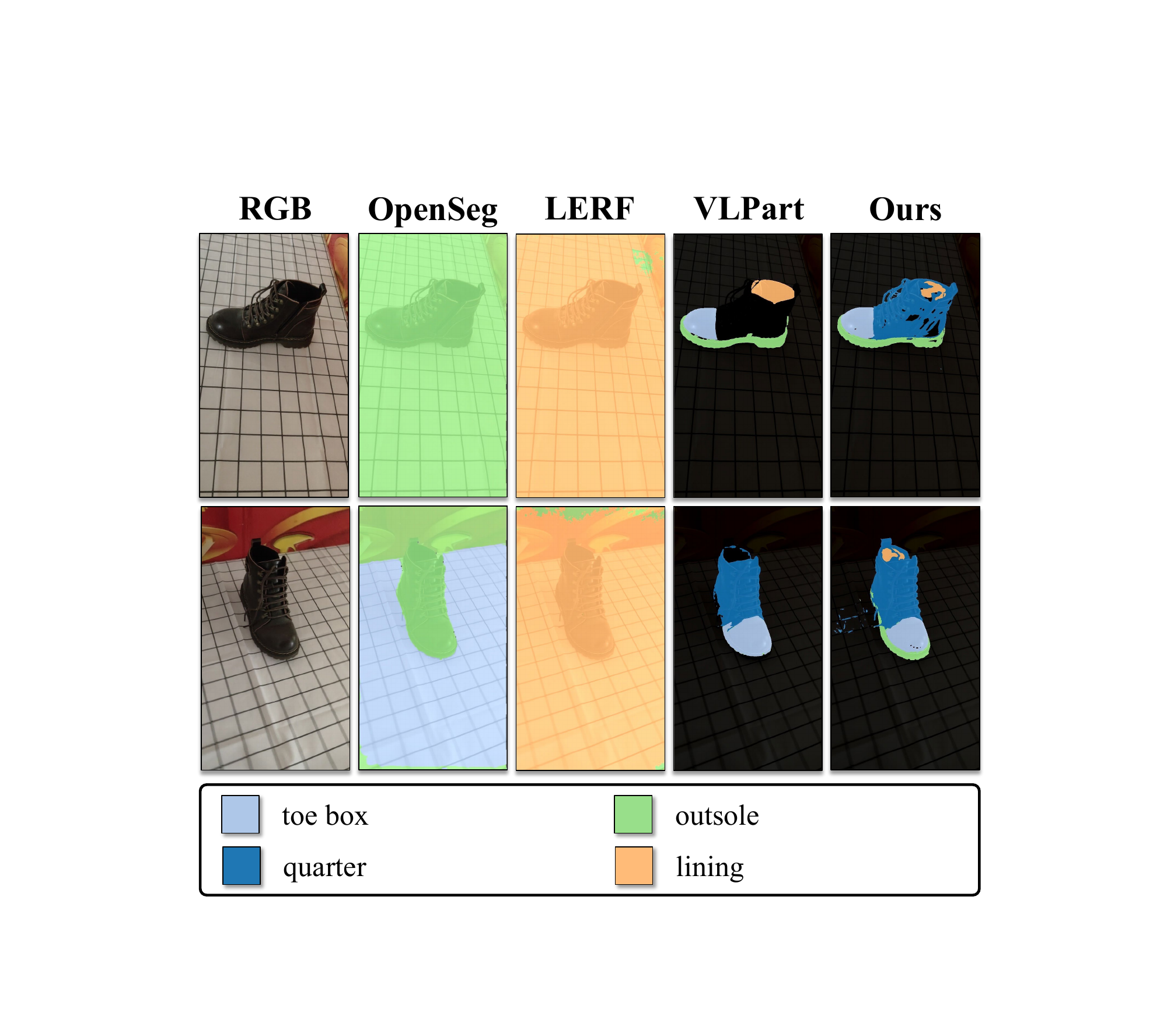}}
    \subfloat[bottle]{\includegraphics[width=0.32\linewidth]{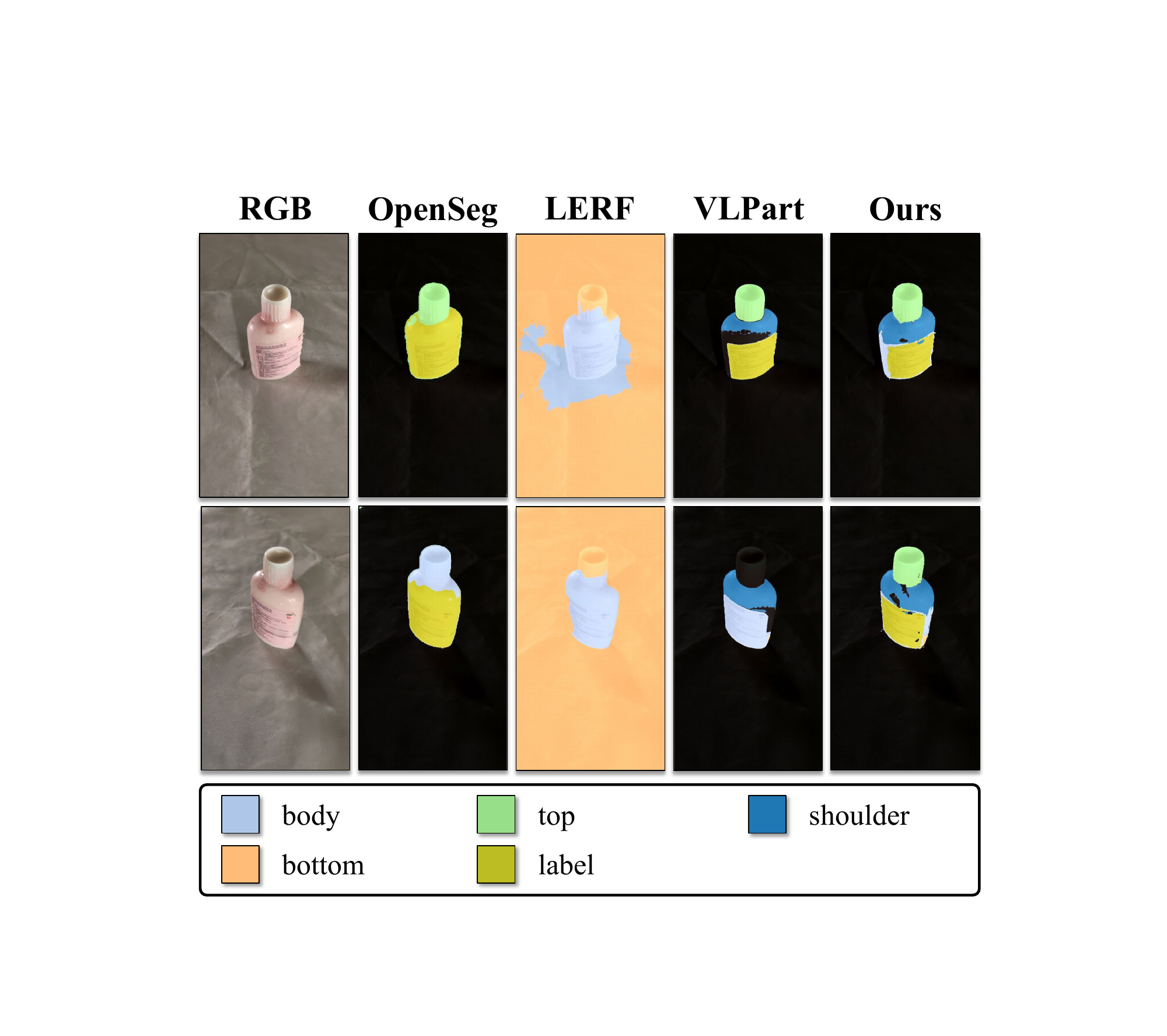}}
    \subfloat[guitar]{\includegraphics[width=0.32\linewidth]{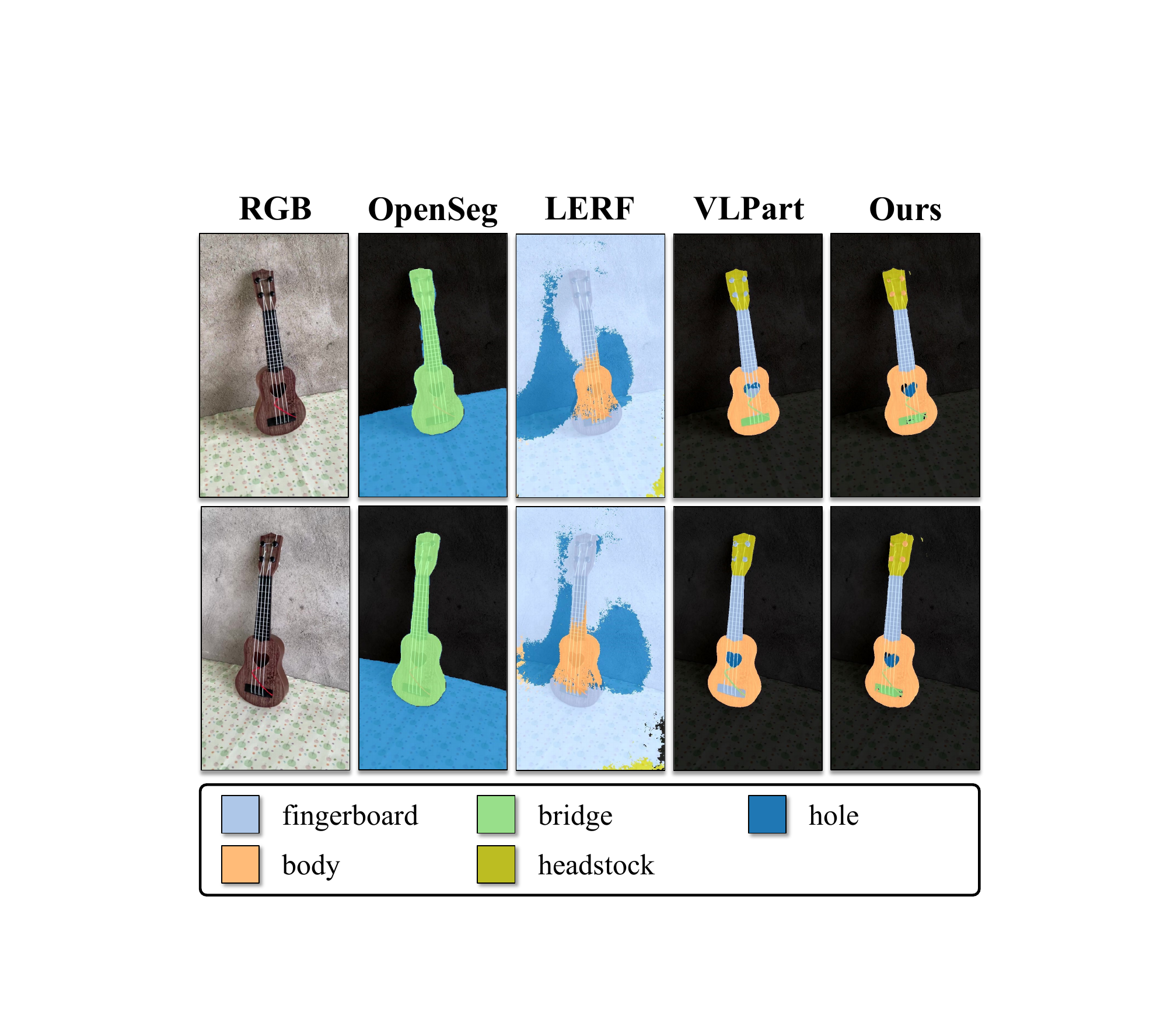}}
\end{center}
\caption{Qualitative comparisons of different methods on the MVImgNet part segmentation task. We choose 6 classes of objects with 3, 4 and 5 parts to show the part segmentation performance.} 
\label{fig:part_seg}
\end{figure*}

\begin{figure*}[!ht]
\begin{center}
    \subfloat[basketball]{\includegraphics[width=0.49\linewidth]{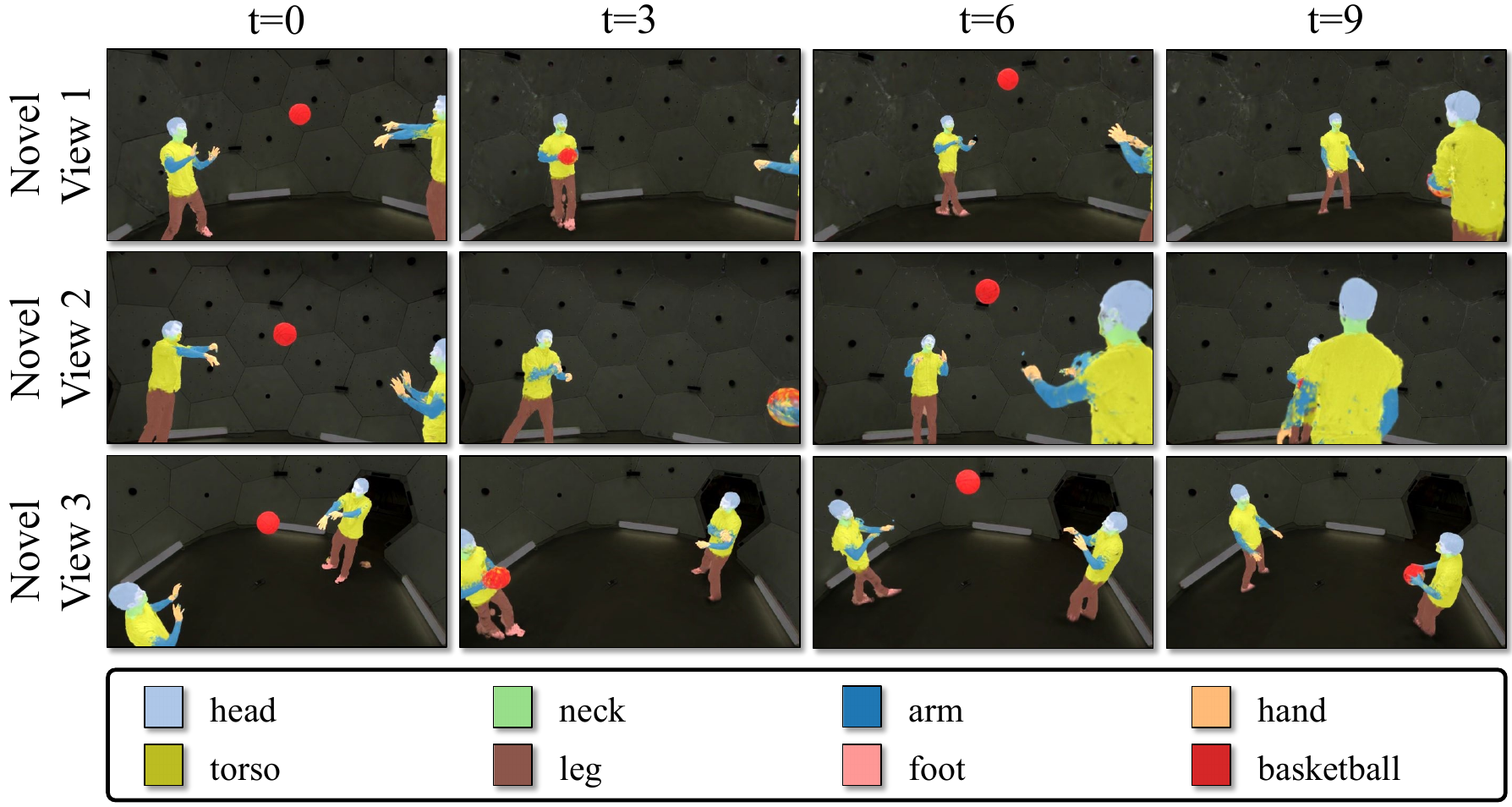}}
    \subfloat[juggle]{\includegraphics[width=0.49\linewidth]{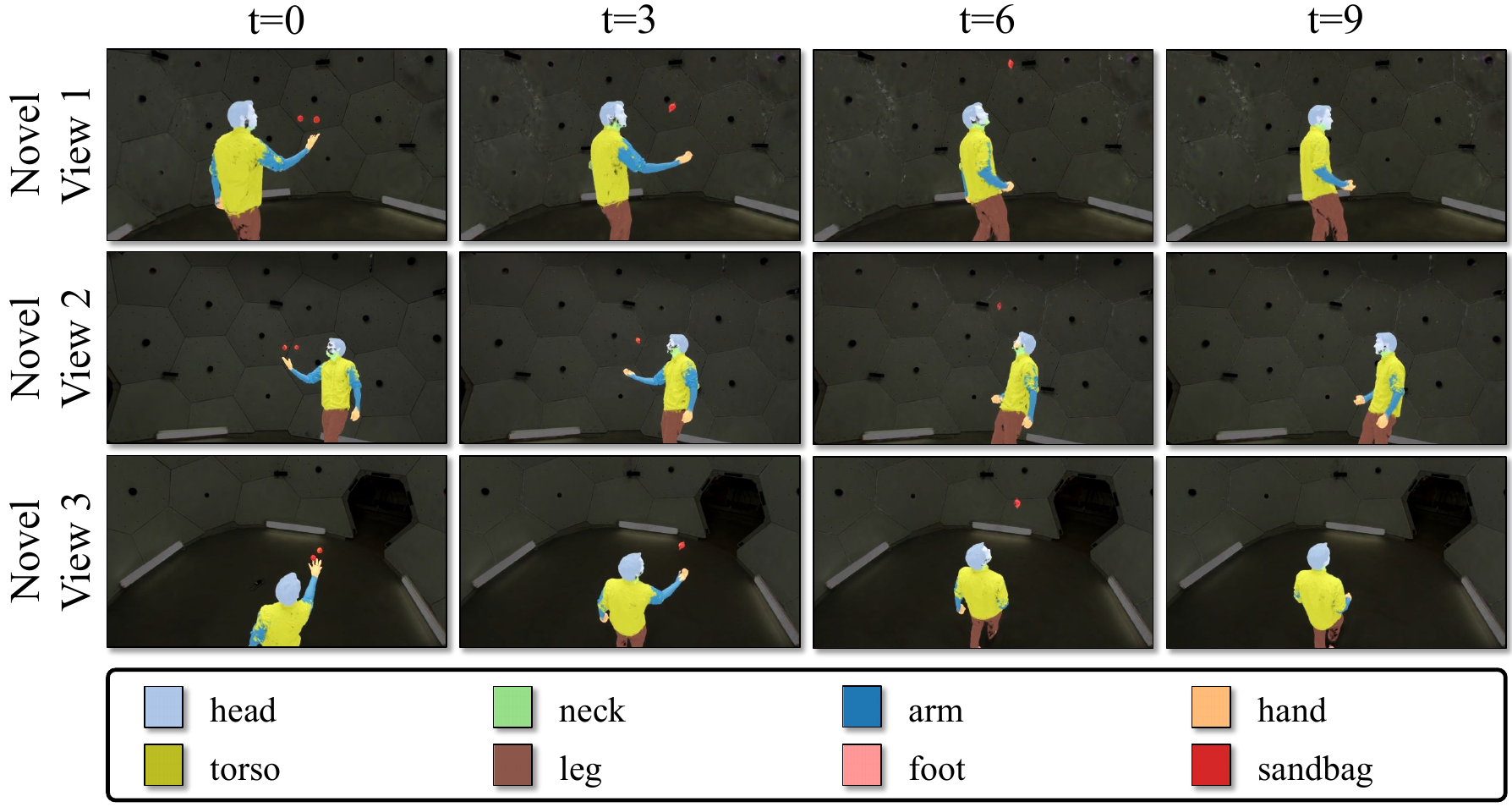}} \\
    \subfloat[softball]{\includegraphics[width=0.49\linewidth]{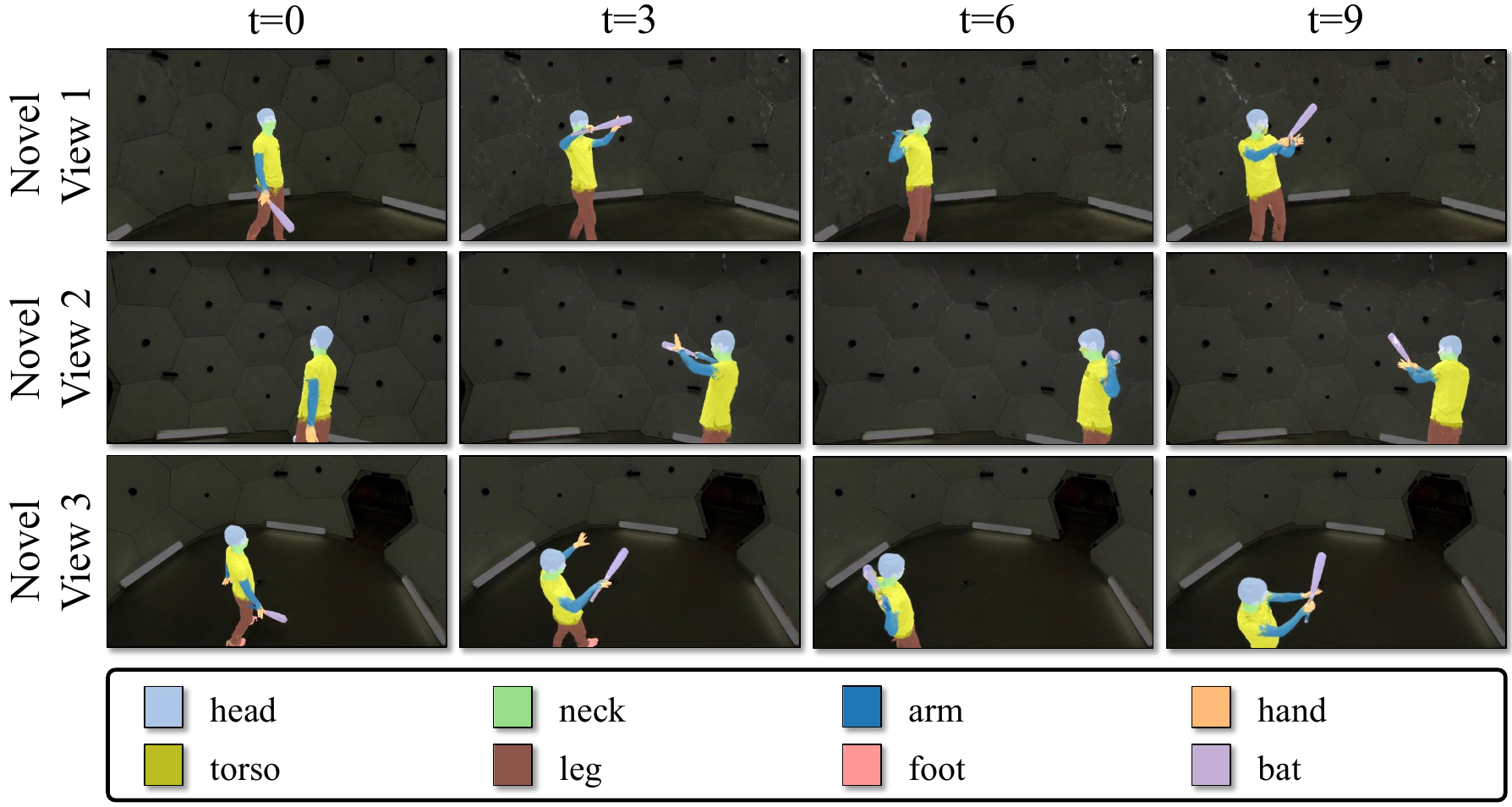}}
    \subfloat[tennis]{\includegraphics[width=0.49\linewidth]{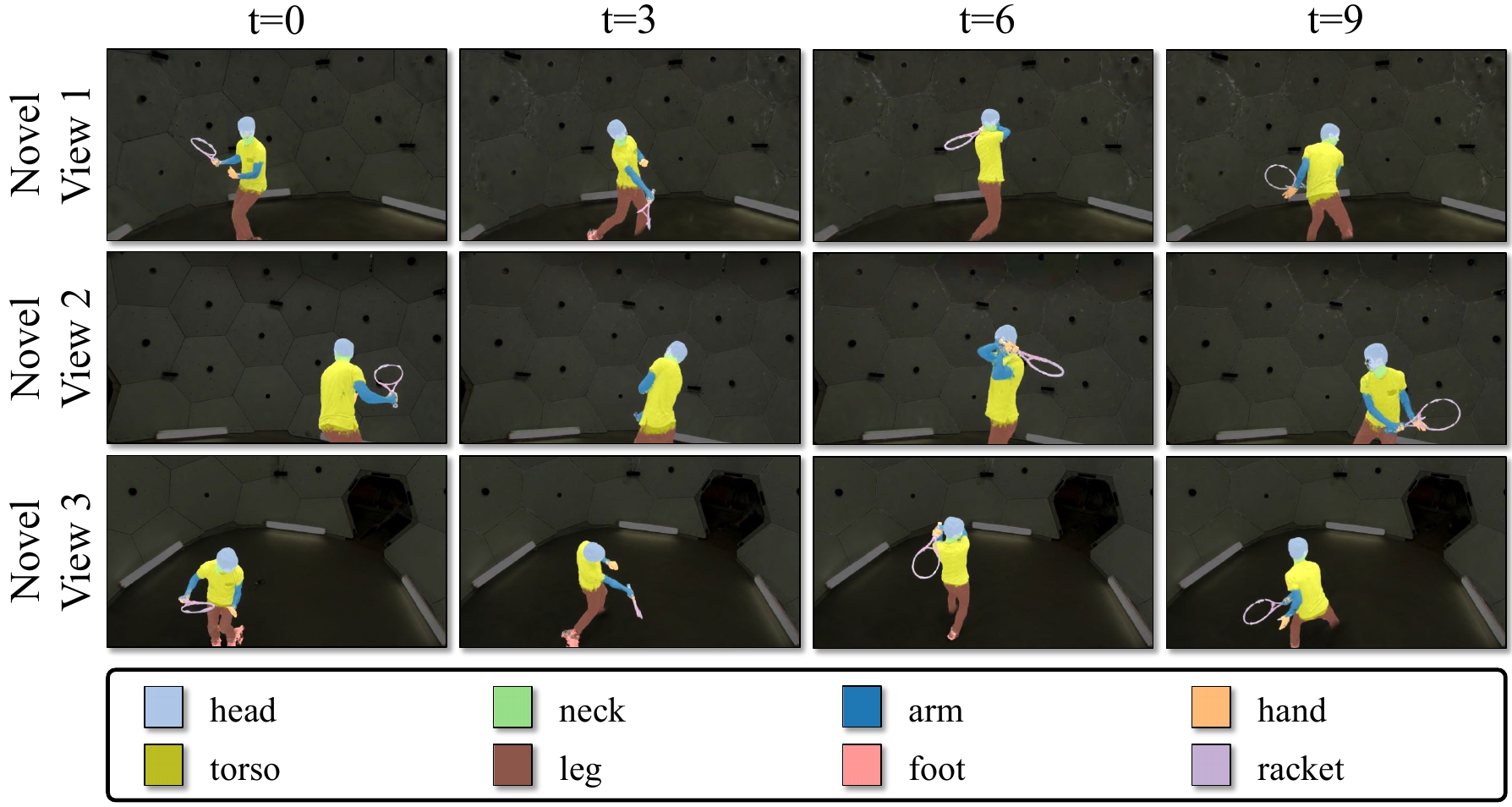}}
\end{center}
\caption{Qualitative results of spatiotemporal tracking on the CMU Panoptic dataset. We choose 4 scenes with humans and dynamic objects to show the tracking performance.} 
\label{fig:dynamic_tracking}
\end{figure*}

We first evaluate our approach on the scene-level semantic segmentation task. We compare our method with several methods, including closed-set and open-vocabulary segmentation methods. We choose both 2D segmentation models and 3D segmentation methods based on NeRF or 3DGS. We report mIoU and mAcc as the metrics of semantic segmentation on the ScanNet-20 benchmark. We compare our method on ScanNet dataset. ScanNet dataset has several canonical classes (wall, floor, table, \etc) with ground truth semantic labels on each posed 2D image. As we only process off-the-shelf 3D Gaussians and do not care about their training process, we do not need to report metrics about rendering quality. The training of 3D Gaussian Splatting follows the official setting. In our method, we extract semantic features from LSeg to conduct 2D projection and 3D network training. Consequently, the result in this section will show how much our method will improve from 2D vision-language models.

Table~\ref{tab:2d_seg} shows the performance of different methods. It can be observed that our \method surpasses all open-vocabulary methods in mIoU and mAcc, and closely approaches the performance of state-of-the-art closed-set methods. This demonstrates that our approach effectively facilitates 3D scene understanding. We can also observe that both our 2D projection (Sec.~\ref{sec:2d_proj}) and our 3D network (Sec.~\ref{sec:3d_network}) surpass the pre-trained LSeg model, even though all of our knowledge comes from it and relies on no ground truth labels. We assume the reason lies in the multi-view information integration, keeping the prediction consistent and mitigating some errors in low-quality views. Moreover, we notice that though the mIou and mAcc of our 3D network are lower than the 2D projection, the 2D and 3D ensemble will further improve our performance. We conjecture that some objects in certain scenes cannot be correctly recognized by 2D models in all views due to their low quality, while the 3D network could recognize them by utilizing geometric details.

Fig.~\ref{fig:sem_seg} shows the segmentation results of open-vocabulary methods based on NeRF and 3DGS on the ScanNet dataset. As shown, LERF and LangSplat exhibit low segmentation accuracy. This is primarily because they utilize multi-scale CLIP features for scene understanding, while CLIP features struggle to align precisely with the scene at the pixel level, making them unsuitable for generating pixel-level semantic segmentation maps. On the other hand, PVLFF, Feature 3DGS, and \method all employ LSeg to achieve open-vocabulary scene understanding. PVLFF performs relatively poorly and also suffers from slow rendering speeds as a NeRF-based method. Feature 3DGS and our method yield very similar performance, with Feature 3DGS even achieving more precise segmentation in certain views. However, Feature 3DGS requires retraining the semantic 3DGS for each scene, which is less efficient and flexible compared to our method.

\subsubsection{Object Localization}
\label{sec:localization}

Subsequently, we evaluated our method on the 3D object localization task across 4 scenes. Note that the LSeg model performed poorly on this task, significantly lagging behind CLIP-based methods like LERF and LangSplat. This likely lies in the fact that LSeg lacks the ability to recognize and segment long-tail objects, making it ineffective in locating objects within this dataset. Therefore, results of other LSeg-based methods are not included in the table. Fortunately, our method is not restricted by the 2D pre-trained model and can utilize SAM as a mask generator, combining it with the CLIP model to achieve versatile projection. Our \method based on SAM+CLIP outperformed in 3 out of 4 scenarios in the segmentation task and achieved the highest average accuracy, demonstrating the effectiveness and flexibility of our method.

\begin{figure}[!t]
\begin{center}
    \includegraphics[width=\linewidth]{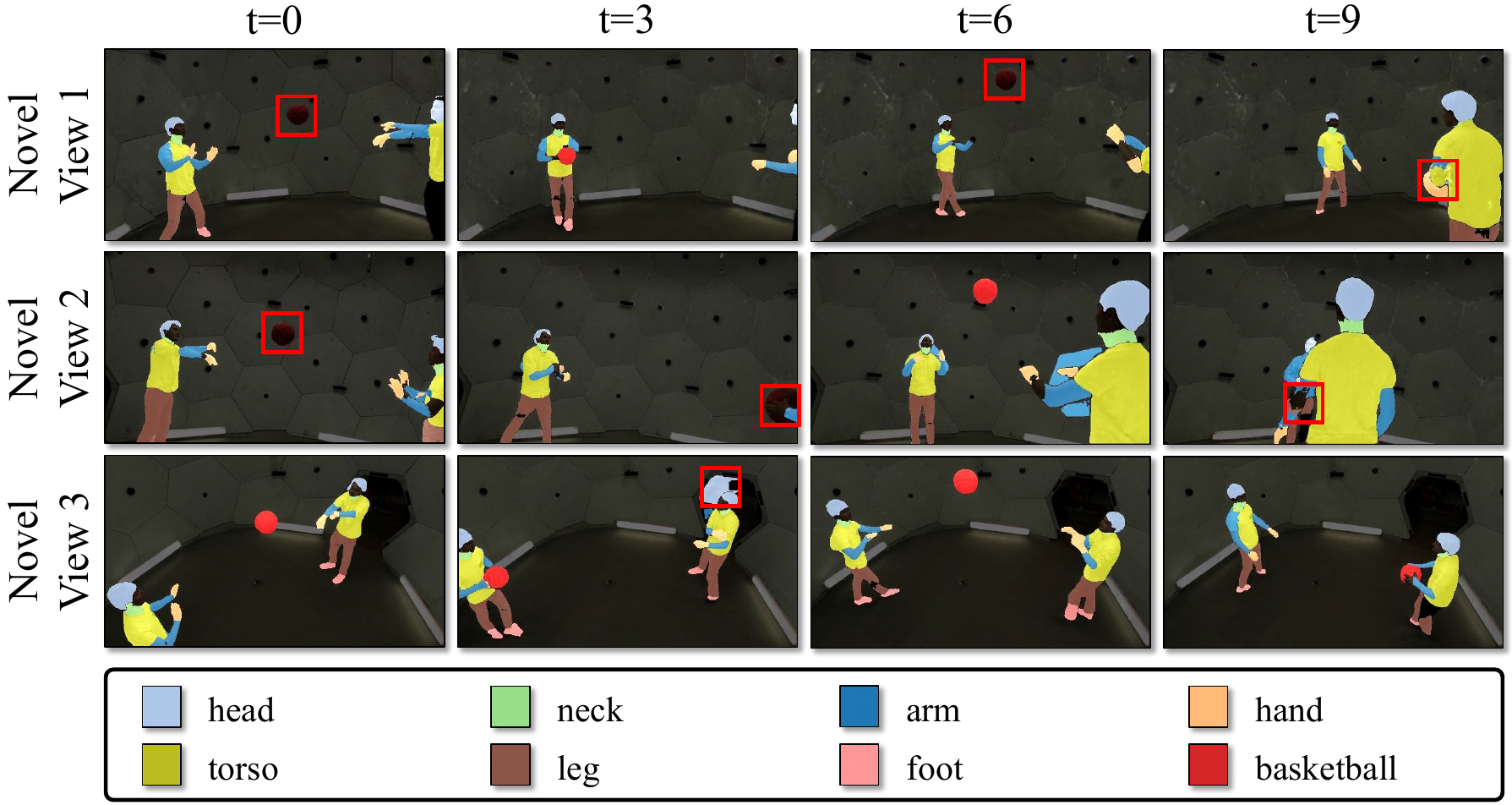}
\end{center}
\caption{Visualization performance of VLPart~\cite{vlpart} on CMU Panoptic dataset. The failure cases are highlighted by red boxes.} 
\label{fig:dynamic_tracking_vlpart}
\end{figure}

\subsubsection{Ablation Study}
\label{sec:ablation}

We conduct ablation studies to figure out the efficiency of our method. As the performance of 2D projection and 3D network is presented in Sec.~\ref{sec:sem_seg}, in this section, we ablate the performance of reducing 3D network input features, Gaussian points and input views. Points in 3D Gaussian Splatting have many features, \ie, coordinates, colors, rotations, scales, and opacities. If we only use coordinates and colors, they degrade to point clouds. Table~\ref{tab:ablation} shows the comparisons of different inputs. We observe that if we reduce the input features, the mIoU and mAcc of our 3D networks will become much lower. This implies that the extra features in 3D Gaussian Splatting are important, and provide more information than RGB point clouds. Table~\ref{tab:ablation} also shows the impact of reducing the number of Gaussian points and the number of input views, although these effects are less significant than reducing the input dimensions of the 3D network. This indicates that our method maintains a certain level of robustness even when there are fewer Gaussian points or input views, without suffering severe performance degradation.

\subsection{Qualitative Evaluations}

\subsubsection{Part Segmentation}

In this section, we show the application of our \method on part segmentation. As OpenSeg cannot tell object parts correctly, here we extract features from VLPart~\cite{vlpart} and use SAM~\cite{sam} to refine the segmentation result. The experiments are conducted on the MVImgNet dataset, which has different types of single objects suitable for part segmentation. The MVImgNet dataset does not have ground truth segmentations, so we compare our methods with other baseline models. Specifically, we compare our method with 2D vision-language models including OpenSeg and VLPart, and LERF~\cite{lerf}, a NeRF-based method that distills knowledge from multi-scale CLIP.

Fig.~\ref{fig:part_seg} shows the qualitative result of part segmentation. From the result, we find that OpenSeg and LERF cannot distinguish object parts correctly, as their capabilities are limited by their training data. By contrast, VLPart is trained on object part segmentation dataset and it can tell parts effectively. However, VLPart cannot keep segmentation consistency across different views, while our \method distills knowledge from VLPart, and can keep high-quality segmentations in all views.

\subsubsection{Spatiotemporal Tracking}

In this section, we show the performance of our \method on spatiotemporal tracking. 3D Gaussian Splatting is originally designed to represent a static scene, while some succeeding works~\cite{dynamic3dgaussians,4dgs,spacetimegaussian} ameliorate it to support 4D dynamic scenes. We follow the work of Dynamic 3D Gaussians~\cite{dynamic3dgaussians} to represent a spatiotemporal scene and use their pre-trained scenes from the CMU Panoptic dataset~\cite{cmupanoptic} to evaluate our tracking performance. 

Fig.~\ref{fig:dynamic_tracking} shows the qualitative result of our spatiotemporal tracking. As the scene often contains 1\~2 humans and a dynamic object, we use VLPart in 2D projection and perform human part segmentation simultaneously. For each dynamic 3D Gaussian, we conduct our method at every frame to avoid looking at future frames. We render some novel views to show our capability of spatial tracking. The results show that our \method can track human parts and objects with a high accuracy between different views and timesteps.

We also tried to segment each image individually from the same viewpoint by VLPart, and the results are shown in the Fig.~\ref{fig:dynamic_tracking_vlpart}. It can be observed that, although VLPart is able to produce fine boundaries, it makes different errors at different viewpoints or timesteps (highlighted in red boxes in the figure). This lack of spatiotemporal consistency in VLPart's results makes it difficult to achieve spatiotemporal tracking. In contrast, our method, which is based on 4D Gaussians as the representation, inherently possesses strong consistency, enabling more accurate tracking.

\begin{figure}[!t]
\begin{center}
    \includegraphics[width=\linewidth]{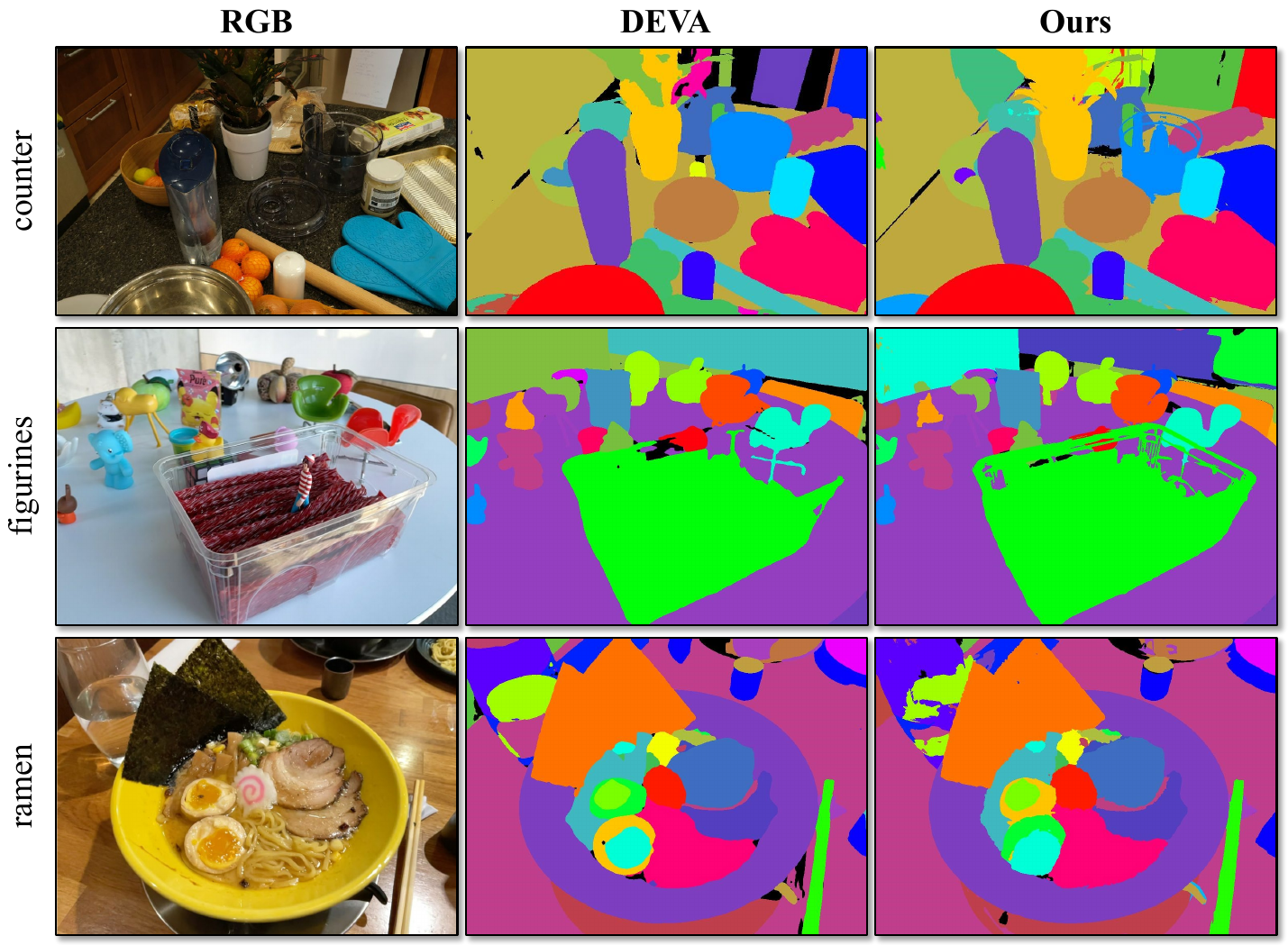}
\end{center}
\caption{Visualization results of instance segmentation results on 3 different scenes. We show the segmentation result of DEVA and our \method. Different colors in the segmentation map denote different instances.}
\label{fig:ins_seg}
\end{figure}

\begin{figure*}[!t]
\begin{center}
    \includegraphics[width=\linewidth]{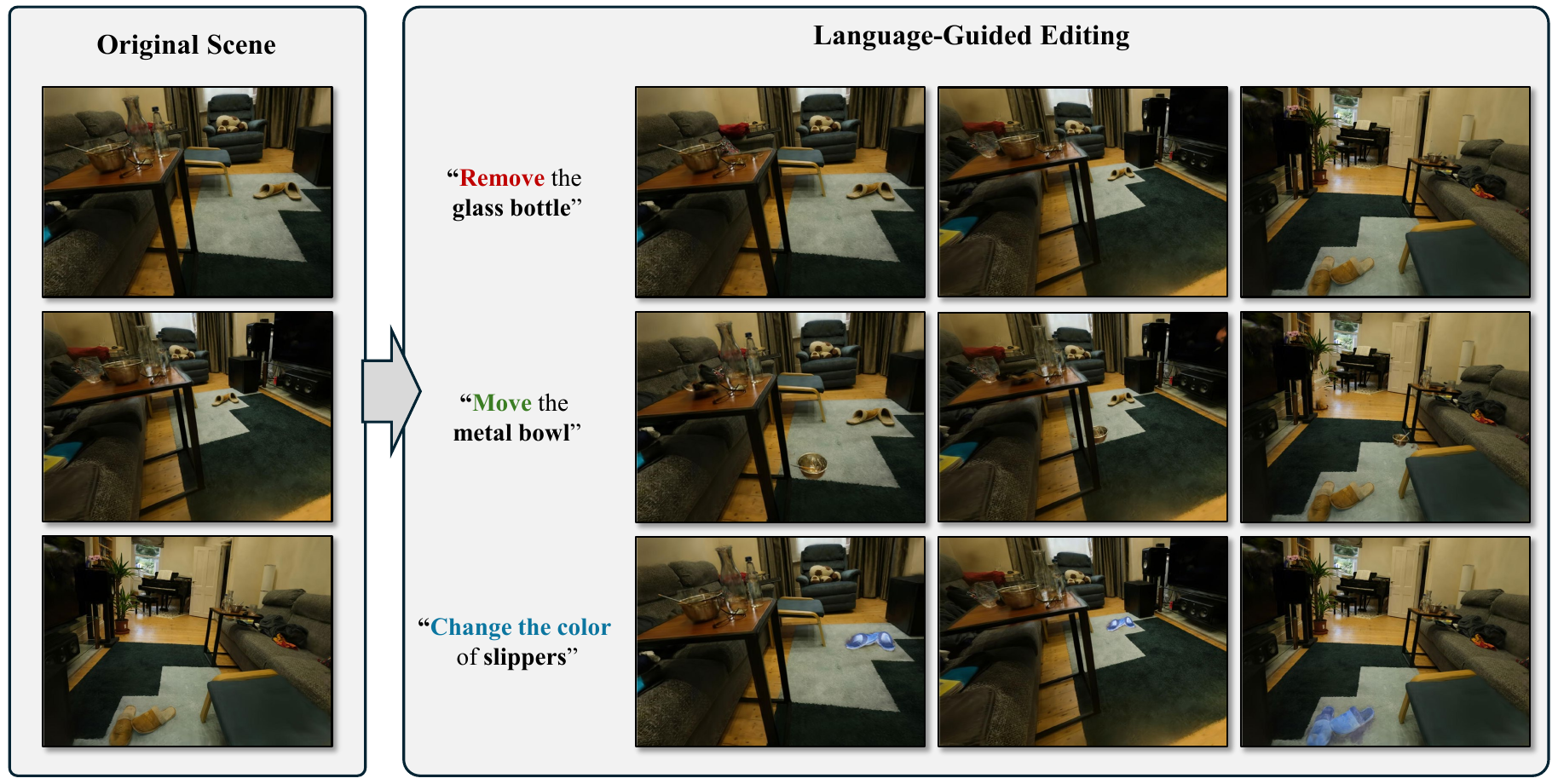}
\end{center}
\caption{Qualitative examples of language-guided editing. We perform object removal, movement, and color change on Mip-NeRF 360 room scene.} 
\label{fig:editing}
\end{figure*}

\subsubsection{Scene-Level Instance Segmentation}

In this section, we will show the performance of \method on scene-level instance segmentation tasks. Some related works~\cite{editing_1, seg3dgs,seg3dgs_2} utilize SAM to generate weakly supervised masks and endow 3DGS with instance segmentation capabilities by redesigning the loss function and training 3DGS. Similarly, our method can leverage the knowledge from SAM to assign instance labels to 3DGS through a single projection. Specifically, following the approach of \cite{editing_1}, we use DEVA~\cite{deva} as the scene-level mask generator. DEVA is capable of assigning a unique scene-level ID to each object in a video, and we project this ID in a one-hot manner directly onto 3DGS. By rendering the semantic channels of 3DGS, instance segmentation maps from any viewpoint can be obtained.

Fig.~\ref{fig:ins_seg} illustrates the visualization results of our method on the instance segmentation task. We choose 3 scenes from different datasets including LERF~\cite{lerf}, 3D-OVS~\cite{3dovs} and Mip-NeRF 360~\cite{mipnerf360}. It can be observed that our method can accurately delineate the boundaries of foreground objects with high precision without any retraining of 3DGS. However, for background objects or those that appear infrequently, our method fails to yield accurate boundaries and instance classifications. The primary reason is that DEVA, as a pre-trained model, cannot ensure the consistency of IDs for the background across different viewpoints. Our method can combine the segmentation results of DEVA from different viewpoints to achieve more consistent 3D instance segmentation results.

\subsubsection{Language-guided Editing}

In this section, we show the application of language-guided editing of our \method. There are several works~\cite{editing_1, editing_2, editing_3, editing_4, editing_5, feature3dgs} that segments 3DGS by SAM and conduct instance editing, but they often require the retraining of 3DGS. Our method can predict semantic embeddings for each 3D Gaussian point, and thus we can choose certain points by language query. We define some canonical operations such as removing, moving and color changing, and employ a language encoder to select the target Gaussians.

Fig.~\ref{fig:editing} shows some qualitative examples of language-guided editing on the room scene in the Mip-NeRF 360 dataset~\cite{mipnerf360}. In these examples, we use CLIP text embeddings of "glass bottle", "metal bowl" and "slippers" to query and choose certain 3D Gaussian points in the scene, and we edit them by modifying their properties such as coordinates, colors, opacities, \etc. From these examples, we observe that the language guidance can accurately select the target object, and thus we can perform various editing operations on the selected 3D Gaussians.

\section{Conclusion}
\label{sec:conclusion}

In this work, we propose \method, a novel approach to open-vocabulary 3D scene understanding via 3D Gaussian Splatting. \method distill knowledge from pre-trained 2D encoders by projecting 2D pixel-level embeddings to 3D Gaussian points. Moreover, we introduce a 3D sparse convolutional network to predict semantic components with the input of RGB Gaussians, thus achieving zero-shot generalization to unseen 3D scenes. We conduct experiments on the ScanNet segmentation benchmark to prove its effectiveness and exhibit downstream applications such as part segmentation, spatiotemporal tracking, instance segmentation, and scene editing. Our work paves the way for real-world applications of 3D Gaussian Splatting, such as embodied agents and augmented reality systems.

Albeit the advantages we have demonstrated, our \method framework does have limitations. The scene understanding performance is bottlenecked by the performance of 2D pre-trained models and off-the-shelf 3D Gaussians. On the one hand, if the 2D pre-trained model completely fails to recognize the scene, \method will fail either. Our proposed 3D semantic network can help lift the performances to some extent. Further, we may reconcile features from multiple 2D pre-trained encoders. On the other hand, if the 3D Gaussians cannot generalize well to a needed novel view, \ie, the 3D Gaussian-based scene representation is weak, \method will not be able to provide robust scene understanding with that novel view as well. This limitation belongs to 3D Gaussian Splatting, as we do not modify any property of 3D Gaussians. Fortunately, 3D Gaussian Splatting is gaining much popularity recently, and we believe the progress made on 3D Gaussian Splatting will improve the performance of our \method.

\bibliographystyle{IEEEtran}
\bibliography{egbib}

\end{document}